\documentclass{article}
\usepackage{ijcai24}

\usepackage{times}
\usepackage{soul}
\usepackage{url}
\usepackage[hidelinks]{hyperref}
\usepackage[utf8]{inputenc}
\usepackage[small]{caption}
\usepackage{graphicx}
\usepackage{amsmath}
\usepackage{amsthm}
\usepackage{booktabs}
\usepackage{algorithm}
\usepackage{algorithmic}
\usepackage[switch]{lineno}
\usepackage{xspace}
\usepackage{multirow}
\usepackage{subfigure}
\usepackage{algorithm}
\usepackage{xspace}
\usepackage{dsfont}
\usepackage{epsfig}
\usepackage{epstopdf}
\usepackage{booktabs}
\usepackage{indentfirst}
\usepackage{amsmath}
\usepackage{amsthm}
\usepackage{amsfonts}
\usepackage{bm}
\usepackage{listings}
\usepackage{tabu}
\usepackage{longtable}
\usepackage{threeparttable}
\usepackage{marvosym}
\usepackage{tabularx}

\usepackage{color}

\usepackage{multirow}
\usepackage{subfigure}
\usepackage{svg} 
\usepackage{subcaption}

\usepackage{dsfont}
\usepackage{epsfig}
\usepackage{epstopdf}
\usepackage{booktabs}

\usepackage{amsmath}
\usepackage{amsthm}
\usepackage{amsfonts}
\usepackage{bm}

\usepackage{longtable}
\usepackage{threeparttable}
\usepackage{marvosym}
\usepackage{tabularx}

\newcommand{\name}{CEGNCDE\xspace}

\newcommand{\pthree}{PeMS03\xspace}
\newcommand{\pfour}{PeMS04\xspace}
\newcommand{\pseven}{PeMS07\xspace}
\newcommand{\peight}{PeMS08\xspace}
\newcommand{\psevenL}{PeMS07 (L)\xspace}
\newcommand{\psevenM}{PeMS07 (M)\xspace}
\newcommand{\upMAE}{2.34\%\xspace}
\newcommand{\upRMSE}{0.97\%\xspace}
\newcommand{\upMAPE}{3.17\%\xspace}

\title{Continuously Evolving Graph Neural Controlled Differential Equations for Traffic Forecasting}

\author{
    Author Name
    \affiliations
    Affiliation
    \emails
    email@example.com
}

\begin{document}

\maketitle

\begin{abstract}
As a crucial technique for developing a smart city, traffic forecasting has become a popular research focus in academic and industrial communities for decades. This task is highly challenging due to complex and dynamic spatial-temporal dependencies in traffic networks. Existing works ignore continuous temporal dependencies and spatial dependencies evolving over time. In this paper, we propose Continuously Evolving Graph Neural Controlled Differential Equations (CEGNCDE) to capture continuous temporal dependencies and spatial dependencies over time simultaneously. Specifically, a continuously evolving graph generator (CEGG) based on NCDE is introduced to generate the spatial dependencies graph that continuously evolves over time from discrete historical observations. Then, a graph neural controlled differential equations (GNCDE) framework is introduced to capture continuous temporal dependencies and spatial dependencies over time simultaneously. Extensive experiments demonstrate that \name outperforms the SOTA methods by average \upMAE relative MAE reduction, \upRMSE relative RMSE reduction, and \upMAPE relative MAPE reduction.
\end{abstract}

\section{Introduction}

\noindent Traffic forecasting, as a crucial technique for developing a smart city and one of the most important parts of intelligent transportation system (ITS)~\cite{ITS} with a great effect on daily life, has become a popular research focus in academic and industrial communities for decades. Accurate traffic forecasting helps citizens save time and costs by avoiding congested roads and peak hours, while also improving the efficiency of the road network and providing reliable guidance for transportation resource scheduling, congestion mitigation, public safety warnings, and daily commuting suggestions.

Traffic forecasting attempts to forecast the future traffic data, e.g., traffic flow and speed, given historical traffic conditions and underlying road networks. This task is highly challenging due to complex and dynamic spatial-temporal dependencies in traffic networks. Deep learning models have been developed to tackle the challenging task of traffic forecasting. As early solutions, sequential models, e.g., recurrent neural networks (RNNs)~\cite{stfgnn} and temporal convolutional networks (TCNs)~\cite{TCN}, are used to capture temporal dependencies. Spatially, these works apply convolutional neural networks (CNNs)~\cite{CNN} to grid-based traffic data to capture spatial dependencies, yet ignore non-Euclidean spatial dependencies~\cite{nonEuc} that are commonly represented using graph structures. Graph neural networks (GNNs)~\cite{GNN} have later been found more suitable for modeling the underlying graph structure of traffic data. Hence, GNN-based methods, in which each node represents a traffic monitor station and each edge indicates a spatial dependency between stations, have been widely explored in traffic forecasting. More recently, spatial-temporal graph neural networks (STGNNs)~\cite{STGNN} have attracted increasing attention for their state-of-the-art (SOTA) performance, by combining GNNs for capturing spatial dependencies with sequential models for capturing temporal dependencies.

Despite the significant success achieved by STGNNs, there are still two limitations. On the one hand, the traffic data used by STGNNs are inherently discrete sequences of observations of a continuous process in the real world~\cite{OnODE}. This means that when using sequence models to capture temporal dependencies, it cannot reflect the influence of all historical traffic conditions on future traffic data. On the other hand, in the real world, spatial dependencies evolve continuously over time, but STGNNs are unable to capture such continuous spatial dependencies.

Neural controlled differential equations (NCDEs)~\cite{NCDE}, considered as the continuous counterpart of RNNs, can model time-series data continuously and capture the influence of all historical information on the forecasted data. STG-NCDE~\cite{GNCDE} introduces NCDE into the traffic forecasting field, using NCDE to capture temporal dependencies and GNN to capture spatial dependencies. However, when using GNN to capture spatial dependencies, STG-NCDE still uses a static graph, which means that it does not support spatial dependencies continuously evolving over time. In addition, STG-NCDE simply uses fully connected graph, ignoring local geographic dependencies and global semantics dependencies.

To address the above-mentioned drawback, we propose Continuously Evolving Graph Neural Controlled Differential Equations (CEGNCDE) for traffic forecasting. To the best of our knowledge, CEGNCDE is the first work that incorporates the continuously evolving graph (CEG) into traffic forecasting, which can not only capture continuous temporal dependencies, but also capture continuous spatial dependencies over time based on the CEG. The main contributions of this paper are outlined as follows:
\begin{itemize}

\item Introduce a continuously evolving graph generator (CEGG) that leverages NCDE and attention mechanism, which can explicitly model spatial dependencies continuously evolving over time. Moreover, we explicitly model the relatively stable components in the spatial dependencies to optimize the CEG.

\item Propose a graph neural controlled differential equations (GNCDE) framework that integrates CEG-based GNN into the forward computation of NCDE to capture continuous temporal dependencies and spatial dependencies over time simultaneously, which can also model both local geographic dependencies and global semantic dependencies via geographic mask and semantic mask.

\item Evaluate CEGNCDE on 6 real-world traffic datasets for traffic data and traffic speed forecasting. The comprehensive experimental results demonstrate the SOTA performance of CEGNCDE, outperforms the best baseline by average \upMAE relative MAE reduction, \upRMSE relative RMSE reduction, and \upMAPE relative MAPE reduction.
\end{itemize}

\section{Related Work}

\subsection{Traffic Forecasting}

\noindent Traffic forecasting has non-trivial effects on our daily life and thus has attracted extensive research attention for decades. Earlier works are usually based on analysis methods, e.g., autoregressive integrated moving average (ARIMA)~\cite{ARIMA}, vector autoregressive (VAR)~\cite{VAR}, and support vector regression (SVR)~\cite{SVR}, which can only capture temporal dependencies, ignoring spatial dependencies. Encouraged by the rapid development of deep learning, RNNs and TCNs are used to capture temporal dependencies, e.g., FC-LSTM~\cite{FC-LSTM}, and CNNs are applied to grid-based traffic data to capture spatial dependencies. These works cannot capture non-Euclidean spatial dependencies dominated by road networks. Towards this problem, GNNs have shown their superior ability. Recently, STGNNs, e.g., STGCN~\cite{stgcn}, ASTGCN~\cite{ASTGCN}, and LSGCN~\cite{lsgcn}, that integrate GNNs to capture spatial dependencies with RNNs or TCNs to model temporal dependencies have shown SOTA performance. However, the performance of STGNNs highly relies on pre-defined graphs constructed based on prior or domain knowledge~\cite{dstcgcn}.

To address this problem, some works, e.g., WaveNet~\cite{WaveNet} and MSTGCN~\cite{MSTGCN}, learn graphs from observations. Due to the learned static graphs, they fail to capture dynamic spatial dependencies. More recently, there are some works, e.g., STSGCN~\cite{stsgcn} and STFGNN~\cite{stfgnn}, that learn dynamic graphs by using learnable node embeddings or dynamic features of nodes, which can model dynamic spatial dependencies. However, these works only focus on dynamics of spatial dependencies, which ignore continuous spatial dependencies that evolve over time.

\subsection{Neural Differential Equations for Time Series}

\noindent Neural differential equations (NDEs) are elegant formulations combining continuous-time differential equations with neural networks~\cite{OnODE}. NODE~\cite{ODE} is the first work to propose that residual networks~\cite{resnet} are continuous simulations of neural differential equations (ODEs), and designs a universal method for constructing neural differential equations. Latent-ODE~\cite{latent} generalizes RNNs to have continuous-time hidden states defined by ODEs, which models the hidden states in RNNs as locally continuous trajectories. GRU-ODE-Bayes~\cite{GRU-ODE-Bayes} and NSDE~\cite{nsde} respectively introduce Bayesian networks~\cite{bayesnet} and Wiener process~\cite{Wiener} to handle occasional observations in the time series. STGODE~\cite{stgode} develops a continuous version of GNN to capture spatial dependencies between multiple time series by replacing residual connections between convolutional layers with neural differential equations.

Although these works have achieved great success in using NODEs to generalize time series models, the hidden states of time series can only be locally continuously modeled, as NODEs take observations from one time step as input~\cite{ODE2}. Inspired by the rough path theory~\cite{paththeory}, NCDEs constructs a general modeling method of neural differential equations that interpolates discrete observations into continuous control paths equations, which can take observations from all time steps as inputs. 

Recently, due to their abilities to continuously model neural networks, NCDEs have been introduced for traffic forecasting. STG-NCDE~\cite{GNCDE} introduces NCDEs to capture continuous temporal dependencies and spatial dependencies that reflect the influence of all historical traffic conditions on future traffic data. However, they still use static graphs, which neglect spatial dependencies evolving over time. In our work, the CEGG module that leverages NCDE and attention mechanism is introduced to explicitly model spatial dependencies continuously evolving over time. 

\section{Preliminaries}
\subsection{Notations and Definitions}

\paragraph{Definition 1: Traffic Data Tensor}
We use ${\bm{X}^{t_1:t_T}} = [{{\bm{X}^{t_1}}, {\bm{X}^{t_2}}, \cdots, {\bm{X}^{t_T}}}] \in \mathbb{R}^{T \times N \times C}$ to denote the traffic data tensor of all nodes at total $T$ time steps, where $C$ is the dimension of the traffic data. We use ${\bm{X}^t} \in {\mathbb{R}^{N \times C}}$ to denote the traffic data at time $t$ of $N$ nodes in the road network and ${\bm{X}_i^t} \in {\mathbb{R}^{C}}$ denotes the values of all the features of node $i$ at time $t$. 

\paragraph{Definition 2: Road Network}
We represent the road network as a continuous graph $\mathcal{G}(t) = (\mathcal{V}, \mathcal{E}, \bm{A}_{\rm{E}}(t))$, where $\mathcal{V}$ is a set of $N$ nodes and $\mathcal{E} \subseteq \mathcal{V} \times \mathcal{V}$ is a set of edges. We use $\bm{A}_{\rm{E}}(t) \in {\mathbb{R}^{N \times N}}$ to denote the continuous adjacency matrix at time $t$.

\subsection{Problem Formalization}
\noindent The aim of traffic forecasting is to forecast the traffic data at the future time given the historical observations. Formally, given the traffic data tensor $[{{\bm{X}^{t_1}}, {\bm{X}^{t_2}}, \cdots, {\bm{X}^{t_T}}}]$ and road network $\mathcal{G}(t)$, our goal is to learn a mapping function $\bm{\mathcal{F}}$ to forecast future $T^\prime$ time steps data based on the previous $T$ time steps observations, which can be written as follows:
\begin{equation}\label{eq:problem_def}
[{{\bm{\hat{X}}^{t_{T+1}}}}, \cdots, {{\bm{\hat{X}}^{t_{T+T^\prime}}}}] = {\mathcal{F}}([{\bm{X}^{t_1}}, \cdots, {\bm{X}^{t_T}}]; \mathcal{G}(t)).
\end{equation}

\subsection{Neural Controlled Differential Equations}

\noindent As the foundation of NCDE, a NODE can be written as follows:
\begin{align} \label{eq:node}
\bm{z}(t) = \bm{z}(t_1)+ \int_{t_1}^{t} {f_\theta}(\bm{z}(s)) ds,
\end{align}where the vector field $\bm{f_\theta}(\bm{z}(s))$ is a neural network parameterized by $\bm{\theta}$ and approximates $\frac{d\bm{z}(s)}{ds}$. We rely on existing ODE solvers, e.g., the explicit Euler method~\cite{Eulermethod} and the 4-th order Runge-Kutta (RK4) method~\cite{RK4}, to solve the integral problem that how to derive the final value $\bm{z}(t)$ from the initial value $\bm{z}(t_1)$. To avoid confusion between the upper bound of integration and the integrand variable in the use of notation, we use $s$ to denote the integrand variable, i.e., time. 

NCDEs generalize RNNs in a continuous manner by creating a control path $\bm{X}(s)$ from observations at all time steps. An NCDE can be written as follows:

\begin{align}\label{eq:ncde}
\bm{z}(t) &= \bm{z}(t_1)+ \int_{t_1}^{t} {f_\lambda}(\bm{z}(s)) d\bm{X}(s)\\
&= \bm{z}(t_1)+ \int_{t_1}^{t} {f_\lambda}(\bm{z}(s)) \frac{d\bm{X}(s)}{ds} ds,\label{eq:ncde2}
\end{align}
where $\bm{X}(s)$ is a natural cubic spline path of observations and $\bm{f_\lambda}(\bm{z}(s))$ is a neural network parameterized by $\bm{\lambda}$. The integral form of Eq.~\eqref{eq:ncde} is the Riemann–Stieltjes integral, whereas the integral form of Eq.~\eqref{eq:node}  is the Riemann integral~\cite{Riemann}. We can reduce NCDEs to the form of Riemann integral written as Eq.~\eqref{eq:ncde2}.

\section{Methodology}
\begin{figure}[t]
    \centering
    \includegraphics[width=0.9\columnwidth]{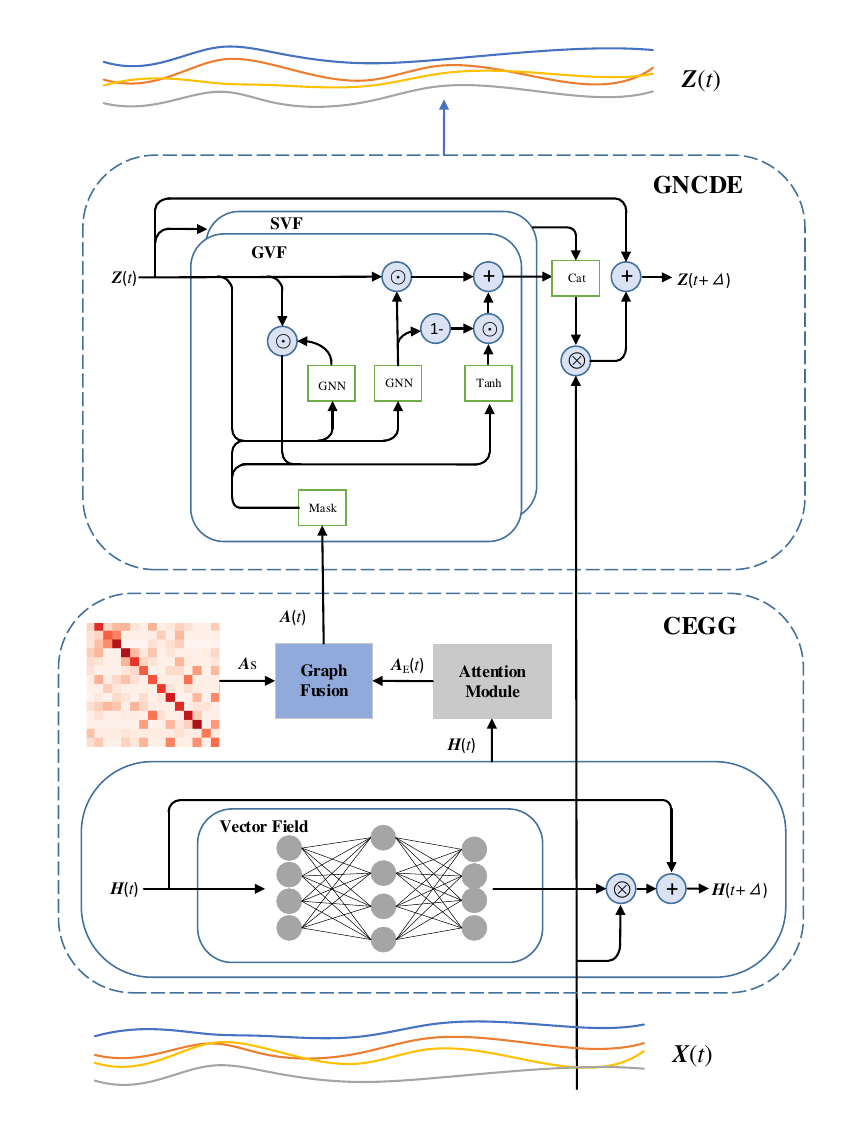}
    \caption{Overall Framework of \name.}
    \label{fig:framework}
\end{figure}

\subsection{Model Overview}
%
 \noindent Capturing continuous temporal dependencies and spatial dependencies evolving over time simultaneously via CEG is challenging for: 1) generating CEG from discrete observations while controlling the scale of memory footprint; and 2) combining CEG-based operations that capture continuous spatial dependencies into operations that capture continuous temporal dependencies. Therefore, we propose CEGNCDE (demonstrated in  Figure~\ref{fig:framework}) to tackle these difficulties.

Specifically, a continuously evolving graph generator (CEGG) based on NCDE is introduced to generate the spatial dependencies graph that continuously evolves over time from discrete historical observations. Since we learn a graph generator through NCDE and attention mechanism, we avoid the huge memory footprint caused by generating a large number of graphs. In addition, we also constructed a static graph in CEGG to explicitly model global stable spatial dependencies. A graph neural controlled differential equations (GNCDE) framework is introduced to capture continuous temporal dependencies and spatial dependencies over time simultaneously, which takes the CEG-based operations that capture spatial dependencies as the vector field function of GNCDE that captures continuous temporal dependencies. Considering that CEG is a fully connected graph, we construct two graph mask methods for CEG, i.e., geographic mask and semantic mask, so that both local geographic dependencies and global semantic dependencies can be modeled. 

\subsection{Continuously Evolving Graph Generator}

 \noindent In a road network, the spatial dependencies between nodes evolve continuously over time. Although existing works~\cite{dynamic} can model dynamic spatial dependencies, such dynamic spatial dependencies are discrete and cannot reflect its continuous evolution over time. In CEGG, we introduce the NCDE and attention mechanism to generate the continuous adjacency matrix $\bm{A}_{E}(t)$ of CEG.

\begin{figure}[t]
    \centering
    \includegraphics[width=0.9\columnwidth]{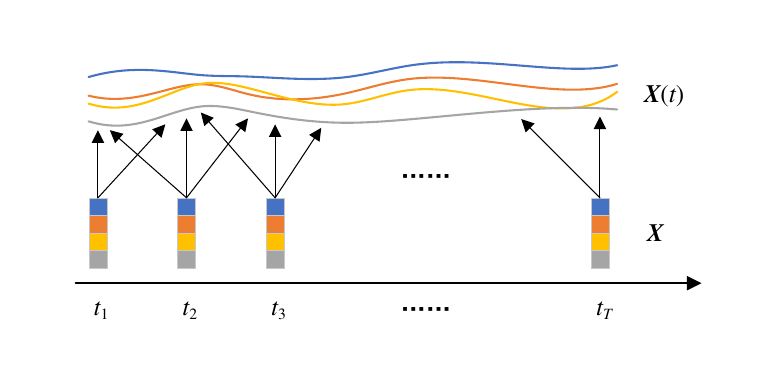}
    \caption{Illustration of Path.}
    \label{fig:path}
\end{figure}

First, we construct an NCDE to generate the hidden state $\bm{H}_{i}(t) \in \mathbb{R}^{d_{\rm{h}}}$ for node $i$ at time $t \in [t_1,t_T]$:
\begin{align}
\bm{H}_{i}(t) &= \bm{H}_{i}(t_1) + \int_{t_1}^{t} {f_\mu}(\bm{H}_{i}(s)) \frac{d\bm{X}_{i}(s)}{ds} ds, \label{eq:type1}
\end{align} where $\bm{X}_{i}(s)$ is a continuous path created from $\bm{X}^{t_1:t_T}$ by the cubic spline interpolation~\cite{path}. As illustrated in  Figure~\ref{fig:path}, the interpolation between adjacent observations is only affected by these two observations and $\bm{X}_{i}(t)={\bm{X}_i^t}, t\in [t_1,t_2,\dots,t_T]$. Since $\bm{X}_{i}(s)$ is third-order differentiable, it is guaranteed that the gradient can be backpropagated when ${\frac{d\bm{X}_{i}(s)}{ds}}$ participates in the integral computation.  

By stacking the hidden states for all nodes, Eq.~\eqref{eq:type1} can be equivalently rewritten as follows using the matrix notation:
\begin{align}
\bm{H}(t) &= \bm{H}(t_1) + \int_{t_1}^{t} {f_\mu}(\bm{H}(s)) \frac{d\bm{X}(s)}{ds} ds, \label{eq:type1-2}
\end{align}where $\bm{H}(t) \in \mathbb{R}^{N \times d_{\rm{h}}}$, $\bm{X}(s)$ is a matrix obtained by stacking continuous paths $\bm{X}_{i}(s)$ for all nodes and $d_{\rm{h}}$ is the dimension of the hidden state.
The initial hidden state $\bm{H}(t_1)$ is created from $ \bm{X}(t_1)$ as follows:
\begin{align}
\bm{H}(t_1) = \sigma(\bm{X}(t_1)\bm{W}_{\rm{H}}), \label{eq:h0}
\end{align}where $\bm{W}_{\rm{H}} \in \mathbb{R}^{C \times d_{\rm{h}}}$ are learnable parameters and $\sigma$ is a rectified linear unit.

 The vector field ${f_\mu}(\bm{H}(s))$  parameterized by $\bm{\theta}_f$ separately processes the hidden state $\bm{H}_{i}(t)$. The definition of ${f_\mu}(\bm{H}(s))$ is as follows:
\begin{align}\begin{split}
{f_\mu}(\bm{H}(s)) &= \psi(\bm{B}_{L})\\
\bm{B}_L &= \mathrm{FC}(\bm{H}(s)),\label{eq:fun_f}
\end{split}\end{align}
where $\mathrm{FC}$ is $L$ fully connected-layers. $\psi$ is dimension conversion unit, which contains a fully-connected layer whose input size is $\mathbb{R}^{N \times d_{\rm{h}}}$ and output size is $\mathbb{R}^{N \times (d_{\rm{h}} \times C)}$ and a conversion unit that converts the output size to $\mathbb{R}^{N \times d_{\rm{h}} \times C}$. ${\mu}$ refers to all the parameters of the vector field ${f_\mu}(\bm{H}(s))$, which independently processes each row of $\bm{H}(t)$ with the $L$ fully connected-layers. Since the dimension of ${\frac{d\bm{X}_{i}(s)}{ds}}$ is $\mathbb{R}^{N \times C}$, the conversion of dimensions is necessary so as to guarantee that the integral result and $\bm{H}(t)$ have consistent dimensions.

\begin{figure}[t]
    \centering
    \includegraphics[width=0.9\columnwidth]{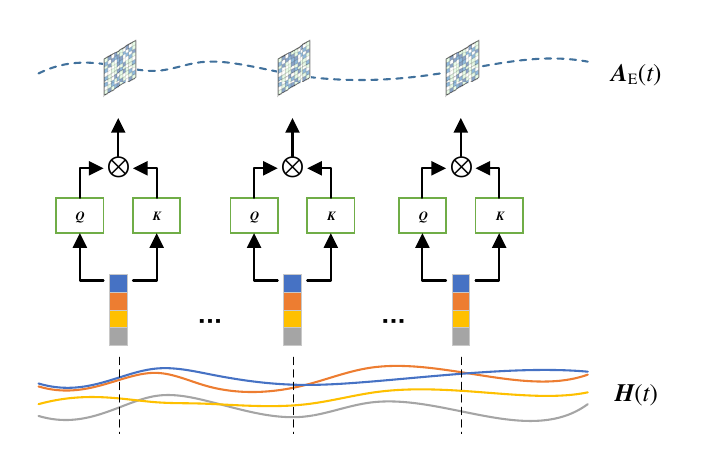}
    \caption{Attention Module of CEGG.}
    \label{fig:attention}
\end{figure}

After obtaining the continuous hidden state $\bm{H}(t)$, we design an attention module (as illustrated in  Figure~\ref{fig:attention}) to model spatial dependencies continuously evolving over time in traffic data. Formally, at time $t$, we first obtain the query and key of attention operations as follows:

\begin{align}\begin{split}
    \bm{Q}(t) &= \bm{H}(t) \bm{W}_{\rm{Q}}\\
    \bm{K}(t) &= \bm{H}(t) \bm{W}_{\rm{K}},
\label{eq:qk}
\end{split}\end{align}where $\bm{W}_{\rm{Q}}, \bm{W}_{\rm{K}} \in \mathbb{R}^{d_{\rm{h}} \times d}$ are learnable parameters and $d$ is the dimension of the query and key matrix. Then, we apply self-attention operations to obtain the spatial dependencies, i.e., attention scores, between nodes at time $t$ as follows:
\begin{equation}\label{eq:At}
    \bm{A}_{\rm{E}}(t)=\frac{\bm{Q}(t)({\bm{K}(t)})^{\top}}{\sqrt{d}}.
 \end{equation}

We noticed that $\bm{A}_{\rm{E}}(t)$ is a continuous function of time, which means that we can explicitly model spatial dependencies continuously evolving over time.

\paragraph{Static Adjacency Matrix}
From a long-term perspective, spatial dependencies between nodes consists of two components. In addition to the component that evolves continuously over time, there is also a component that is relatively stable, e.g., spatial dependencies between two adjacent nodes connected by a road, which reflect their intrinsic correlation. Therefore, we design a learnable static graph to model the stable static spatial dependencies. Specifically, we introduce node embedding~\cite{nodeembedding} to generate a static adjacency matrix $\bm{A}_{\rm{S}}$ in a learnable way. 

We first randomly initialize the learnable embeddings $\bm{E} \in \mathbb{R}^{N \times d_{\rm{e}}}$ for all nodes to represent the static characteristics of nodes that do not change over time, where $d_{\rm{e}}$ denotes the dimension of the node embedding. Then, we multiply $\bm{E}$ and the transpose of $\bm{E}$ to get the static spatial dependencies between nodes. The form is as follows: 
\begin{equation}\label{eq:As}
    \bm{A}_{\rm{S}}=Softmax(\sigma(\bm{E}\bm{E}^{\top})),
 \end{equation}
where the Softmax function is used to normalize the adaptive adjacency matrix and the rectified linear unit $\sigma$ is used to eliminate weak connections.

Through the above two ways of obtaining spatial dependencies, we can obtain not only a stable static adjacency matrix $\bm{A}_{\rm{S}}$, but also a continuously evolving adjacency matrix $\bm{A}_{\rm{E}}(t)$. Then, we fuse the two adjacency matrices in a weighted way to obtain the optimal CEG at time $t$. The form is as follows: 
\begin{equation}\label{eq:A}
    \bm{A}(t)=(1-\beta)\bm{A}_{\rm{E}}(t) + \beta\bm{A}_{\rm{S}},
 \end{equation}where fusion coefficient $\beta \in [0, 1]$ is a hyperparameter and $\bm{A}(t)$ is the adjacency matrix of the optimal CEG. 
 
\subsection{Graph Neural Controlled Differential Equations}

\noindent We introduce graph neural controlled differential equations (GNCDE) framework to capture continuous temporal dependencies and
spatial dependencies over time simultaneously. The GNCDE can be formalized as follows:
\begin{align}\begin{split}
\bm{Z}(t) = \bm{Z}(t_1) + \int_{t_1}^{t} {g_\xi}(\bm{Z}(s);\bm{A}(s)) \frac{d\bm{X}(s)}{ds} ds,\label{eq:type2}
\end{split}\end{align}where $\bm{Z}(t) \in \mathbb{R}^{N \times d_{\rm{z}}}$ is spatial-temporal representations for all the nodes that aggregates continuous temporal and spatial dependencies over the interval $[t_1,t]$ and $d_{\rm{z}}$ is the dimension of the spatial-temporal representation.  
The initial hidden state $\bm{Z}(t_1)$ is created from $ \bm{X}(t_1)$ as follows:
\begin{align}
\bm{Z}(t_1) = \sigma(\bm{X}(t_1)\bm{W}_{\rm{Z}}), \label{eq:z0}
\end{align}where $\bm{W}_{\rm{Z}} \in \mathbb{R}^{C \times d_{\rm{z}}}$ are learnable parameters and $\sigma$ is a rectified linear unit.
\paragraph{Graph Mask}
We introduce two graph mask methods to optimize $\bm{A}(t)$. The CEG resulting from the above method is the fully connected graph, which contains the spatial dependencies between all nodes. However, in the real world, only a few spatial dependencies between nodes are critical. Therefore, we introduce two graph mask matrices $\bm{M}^{\rm{geo}}$ and $\bm{M}^{\rm{sem}}$ to  capture local geographic dependencies and global semantic dependencies in traffic data, respectively.

We give the definition of the geographic mask matrix $\bm{M}^{\rm{geo}}$. The weight is 1 when the geographic distance between node pairs is less than a threshold, and 0 otherwise. 

We compute the similarity of all node pairs, which is implemented by calculating the Dynamic Time Warping (DTW)~\cite{DTW} distance of two historical traffic data sequences. For each node, its semantic neighbors are defined as the $K$ nodes with the highest similarity values. We set the weights between pairs of semantic neighbor nodes to 1, otherwise to 0, to construct the semantic mask matrix $\bm{M}^{\rm{sem}}$.

The key of GNCDE is how to design the vector field function ${g_\xi}(\bm{Z}(s);\bm{A}(s))$ parameterized by $\bm{\xi}$, which captures spatial dependencies by GNN based on the continuous adjacency matrix $\bm{A}(t)$. Since NCDEs are the continuous version of RNNs, we can refer to GRU~\cite{GRU} when designing vector field function ${g_\xi}(\bm{Z}(s);\bm{A}(s))$. Following prior works~\cite{STGNN}, we replace the MLP layers in GRU with our graph convolution layers as follows:

\begin{equation}
\begin{aligned}\label{eq:gru}
& \bm{A}(t) = Softmax(\sigma(\bm{A}(t) \otimes \bm{M}) \\
&\bm{z}(t) =\sigma(\bm{A}(t) \bm{Z}(t) \bm{W}_{\rm{z}} + \bm{b}_{\rm{z}}) \\
&\bm{r}(t) =\sigma(\bm{A}(t) \bm{Z}(t) \bm{W}_{\rm{r}} + \bm{b}_{\rm{r}}) \\
&\bm{\hat{h}}(t) = tanh(\bm{A}(t) (\bm{r}(t) \odot \bm{Z}(t)) \bm{W}_{\rm{\hat{h}}} + \bm{b}_{\rm{\hat{h}}}) \\
&\bm{h}(t) =\bm{z}(t) \odot \bm{Z}(t) + (1 - \bm{z}(t)) \odot \bm{\hat{h}}(t),
\end{aligned}
\end{equation}where $\bm{M}$ is geographic mask matrix $\bm{M}^{\rm{geo}}$ or semantic mask matrix $\bm{M}^{\rm{sem}}$. By processing two different mask methods separately, we can obtain $\bm{h}^{\rm{geo}}(t)$ when $\bm{M}=\bm{M}^{\rm{geo}}$, and $\bm{h}^{\rm{geo}}(t)$ when $\bm{M}=\bm{M}^{\rm{sem}}$. For better distinction, we refer to these two operation processes based on different mask methods as geographic vector field (GVF) and semantic vector field (SVF), respectively. Then, We concatenate them and convert dimensions as follows:
.
\begin{equation}\label{eq:hh}
{g_\xi}(\bm{Z}(t);\bm{A}(t)) = \psi(\mathrm{Cat}(\bm{h}^{\rm{geo}}(t),\bm{h}^{\rm{sem}}(t))),
\end{equation}where $\psi$ is dimension conversion unit, which contains a fully-connected layer whose input size is $\mathbb{R}^{N \times d_{\rm{z}}}$ and output size is $\mathbb{R}^{N \times (d_{\rm{z}} \times C)}$ and a conversion unit that converts the output size to $\mathbb{R}^{N \times d_{\rm{z}} \times C}$. Since the dimension of ${\frac{d\bm{X}_{i}(s)}{ds}}$ is $\mathbb{R}^{N \times C}$, the conversion of dimensions is necessary so as to guarantee that the integral result and $\bm{Z}(t)$ have  consistent dimensions. 

\subsection{Output Module}

 \noindent We take the final spatial-temporal representations $\bm{Z}(t_T)$ to forecast traffic data of all nodes for the next $T^\prime$ time steps by a linear transformation, which decreases the time consumption and avoids cumulative error caused by sequential forecasting. The form is as follows: 
\begin{equation}\label{eq:out}
    \bm{\hat{\mathcal{X}}} = \mathrm{Linear}(\bm{Z}(t_T)),
\end{equation}where $\bm{\hat{\mathcal{X}}} \in \mathbb{R}^{T' \times N \times C}$ is $T^\prime$ time steps forecasting results.

 Since Eq.~\eqref{eq:type1-2} and Eq.~\eqref{eq:type2} always have common integration variables and integration intervals, we can implement them together. To this end, we define the augmented ODE as follows:
\begin{align}
\frac{d}{dt}{\begin{bmatrix}
  \bm{Z}(t) \\
  \bm{H}(t) \\
  \end{bmatrix}\!} = {\begin{bmatrix}
  {g_\xi}(\bm{Z}(s);\bm{A}(s)) \frac{d\bm{X}(s)}{ds} \\
  {f_\mu}(\bm{H}(s)) \frac{d\bm{X}(s)}{ds}\\
  \end{bmatrix}\!}.
\end{align}

In this way, we only need one ODE solver to compute two NCDEs simultaneously. Moreover, this augmented ODE can guarantee that, in the process of capturing spatial dependencies continuously over time, there will always be CEGs generated at corresponding time to support it.

The loss function in this work is $L^1$ loss function:
\begin{align}
\mathcal{L} = \sum\limits_{t_{T+1}}^{t_{T+T^{\prime}}} {\left\|\bm{X}^t - \hat{\bm{X}^t}\right\|},\label{eq:loss}
\end{align}where $\bm{X}^t$ and ${\bm{\hat{X}}^t}$ are the ground truth and the forecasting results.

\section{Experiments}

\subsection{Datasets}
\begin{table}[!ht]
  \centering
  \caption{Datasets Description.}
  \resizebox{\columnwidth}{!}{
  \small{
    \begin{tabular}{l|ccc}
    \toprule
    Datasets & \#Nodes & \#Time Steps  & \#Time Range \\ 
    \midrule
    \pthree & 358      & 26,208   & 09/01/2018-11/30/2018 \\
    \pfour & 307     & 16,992   & 01/01/2018-02/28/2018 \\
    \pseven & 883      & 28,224   & 05/01/2017-08/31/2017 \\
    \psevenM & 228     & 12,672   & 05/01/2012-06/30/2012 \\
    \psevenL & 1026    & 12,672   & 05/01/2012-06/30/2012\\
    \peight & 170      & 17,856  & 07/01/2016-08/31/2016 \\
    \bottomrule
    \end{tabular}%
    }}
  \label{tab:data_detail}%
\end{table}%

\begin{table*}[t]
    \centering
    \caption{Forecasting error on \pthree, \pfour, \pseven, and \peight. Bold denotes the best performance of each metric and underline for the second best.}
    \setlength{\tabcolsep}{4pt}
    \small
    {
    \begin{tabular}{l lll l lll l lll l lll}
        \hline
        \multirow{2}{*}{Model}  & \multicolumn{3}{c}{\pthree}    && \multicolumn{3}{c}{\pfour}      && \multicolumn{3}{c}{\pseven}      && \multicolumn{3}{c}{\peight}\\\cline{2-4} \cline{6-8} \cline{10-12} \cline{14-16}
                                & MAE & RMSE & MAPE             && MAE & RMSE & MAPE               && MAE & RMSE & MAPE               && MAE & RMSE & MAPE               \\ \hline
        HA                      & 31.58  & 52.39  & 33.78         && 38.03  & 59.24  & 27.88           && 45.12  & 65.64  & 24.51           && 34.86  & 59.24  & 27.88           \\       
        ARIMA                   & 35.41  & 47.59  & 33.78         && 33.73  & 48.80  & 24.18           && 38.17  & 59.27  & 19.46           && 31.09  & 44.32  & 22.73  \\        
        VAR                     & 23.65  & 38.26  & 24.51         && 24.54  & 38.61  & 17.24           && 50.22  & 75.63  & 32.22           && 19.19  & 29.81  & 13.10           \\                
        FC-LSTM (NeurIPS, 2006)                 & 21.33  & 35.11  & 23.33         && 26.77  & 40.65  & 18.23           && 29.98  & 45.94  & 13.20           && 23.09  & 35.17  & 14.99           \\ 
       
        \hline          
        STGCN (ICDM, 2020)                   & 17.55  & 30.42  & 17.34         && 21.16  & 34.89  & 13.83           && 25.33  & 39.34  & 11.21           && 17.50  & 27.09  & 11.29           \\         
         WaveNet (IJCAI, 2019)            & 19.12  & 32.77  & 18.89         && 24.89  & 39.66  & 17.29           && 26.39  & 41.50  & 11.97           && 18.28  & 30.05  & 12.15           \\ 
        ASTGCN (AAAI, 2019)             & 17.34  & 29.56  & 17.21         && 22.93  & 35.22  & 16.56           && 24.01  & 37.87  & 10.73           && 18.25  & 28.06  & 11.64           \\ 
        MSTGCN (TNSRE, 2019)               & 19.54  & 31.93  & 23.86         && 23.96  & 37.21  & 14.33           && 29.00  & 43.73  & 14.30           && 19.00  & 29.15  & 12.38           \\       
        LSGCN (IJCAI, 2020)                   & 17.94  & 29.85  & 16.98         && 21.53  & 33.86  & 13.18           && 27.31  & 41.46  & 11.98           && 17.73  & 26.76  & 11.20           \\       
        STSGCN (AAAI, 2020)                 & 17.48  & 29.21  & 16.78         && 21.19  & 33.65  & 13.90           && 24.26  & 39.03  & 10.21           && 17.13  & 26.80  & 10.96           \\        
        STFGNN (AAAI, 2021)                  & 16.77  & 28.34  & 16.30         && 20.48  & 32.51  & 16.77           && 23.46  & 36.60  &  9.21           && 16.94  & 26.25  & 10.60           \\        
        
        \hline
        STGODE (KDD, 2021)                   & 16.50  & {27.84 } & 16.69         && 20.84  & 32.82  & 13.77           && 22.59  & 37.54  & 10.14   && 16.81  & 25.97  & 10.62           \\
        STG-NCDE (AAAI, 2022)      & \underline{15.57 } & \underline{27.09 }    &   \underline{15.06 }  && \underline{19.21 }    & \underline{31.09 }    &  \underline{12.76 }    && \underline{20.53 } & \underline{33.84 } & \underline{8.80 }     && \underline{15.45 } & \underline{24.81 } &  \underline{9.92}  \\
        \hline
        \textbf{CEGNCDE (Ours)}  & \textbf{15.21}    & \textbf{26.93}    &  \textbf{14.85}    && \textbf{18.97} & \textbf{30.99} & \textbf{12.57}     && \textbf{20.36} & \textbf{33.69} &  \textbf{8.60}  && \textbf{15.12} & \textbf{24.56} &  \textbf{8.89}  \\
        
        \hline
    \end{tabular}}
    
    \label{tab:main_exp}
\end{table*}

\begin{table}[!ht]
    \centering
    \caption{Forecasting error on \psevenM and \psevenL. Bold denotes the best performance of each metric and underline for the second best.}
    \setlength{\tabcolsep}{2pt}
    \small
    \scalebox{1}{
    \begin{tabular}{l llllll l lll}
        \hline
        \multirow{2}{*}{Model} & \multicolumn{3}{c}{\psevenM}   && \multicolumn{3}{c}{\psevenL}     \\\cline{2-4} \cline{6-8}
                                & MAE  & RMSE  & MAPE           && MAE  & RMSE  & MAPE               \\ \hline
        HA                      & 4.59 &  8.63 & 14.35         && 4.84 &  9.03 & 14.90          \\       
        ARIMA                   & 7.27 & 13.20 & 15.38         && 7.51 & 12.39 & 15.83          \\        
        VAR                    & 4.25 &  7.61 & 10.28         && 4.45 &  8.09 & 11.62          \\                     
        FC-LSTM                 & 4.16 &  7.51 & 10.10         && 4.66 &  8.20 & 11.69          \\                     
        \hline                   
        STGCN                & 3.86 &  6.79 & 10.06         && 3.89 &  6.83 & 10.09          \\             
         WaveNet            & 3.19 &  6.24 &  8.02         && 3.75 &  7.09 &  9.41          \\       
        ASTGCN              & 3.14 &  6.18 &  8.12         && 3.51 &  6.81 &  9.24          \\       
        MSTGCN                  & 3.54 &  6.14 &  9.00         && 3.58 &  6.43 &  9.01          \\            
        LSGCN                 & 3.05 &  5.98 &  7.62         && 3.49 &  6.55 &  8.77           \\       
        STSGCN                & 3.01 &  5.93 &  7.55         && 3.61 &  6.88 &  9.13           \\        
        STFGNN                & 2.90 & 5.79 & 7.23           && 2.99 & 5.91 & 7.69           \\        
        \hline
        STGODE                 & 2.97 & 5.66 & 7.36          && 3.22 & 5.98 & 7.94         \\
        {STG-NCDE}           & \underline{2.68}    & \underline{5.39}    &  \underline{6.76}  && \underline{2.87} & \underline{5.76} &  \underline{7.31} \\
        \hline 
        \textbf{\name (Ours)}           & \textbf{2.58}    & \textbf{5.25}    & \textbf{6.70}  && \textbf{2.76} & \textbf{5.71} &  \textbf{7.12} \\
        \hline
    \end{tabular}}
    
    \label{tab:main_exp_2}
\end{table}
 \noindent We evaluate the performance of \name on 6 real-world traffic datasets, including PeMS03/4/7/7 (L)/7 (M)/8 provided by California Performance of Transportation (PeMS) in real-time every 30 seconds. To be consistent with most existing studies, we split the 6 real-world traffic datasets chronologically into training, validation, and test sets in a 6:2:2 ratio. We fill the missing data by linear interpolation and aggregate data into 5-minute intervals, resulting in 12 time steps per hour. In addition, we normalize all datasets using the Z-Score normalization method. The detailed dataset statistics are summarized in Table ~\ref{tab:data_detail}.
 
\subsection{Experimental Settings}
 \noindent All experiments are trained on an NVIDIA GeForce RTX 3080 Ti GPU card using Adam optimizer~\cite{adam} with an initial learning rate of 0.001. We implement \name with PyTorch 1.12.0. The depth of fully connected-layers $L$ in ${f_\mu}(\bm{H}(s))$ is set to 3. Both of hidden dimensions, i.e., $d_{\rm{h}}$ and $d_{\rm{z}}$, are set to 64 and the fusion coefficient $\beta$  is set to 0.5. The batch size is 64 and the training epoch is 200. An early stop strategy with a patience of 25 iterations on the vaidation sets is used. For stability, every experiment is repeated 5 times using 5 varying random seeds (1 to 5) and the means of the 5 repeats are reported as the final results. The source code of \name is released online\footnote{https://anonymous.4open.science/r/CEGNCDE}. 

 We use three metrics in the experiments: (1) Mean Absolute Error (MAE), (2) Mean Absolute Percentage Error (MAPE), and (3) Root Mean Squared Error (RMSE). Missing values are excluded when calculating these metrics. 


\begin{table}[!ht]
    \centering
    \caption{Ablation Study on \pfour and \peight. Bold denotes the best performance of each metric.}
    \setlength{\tabcolsep}{2pt}
    \small
   {
    \begin{tabular}{lllllll l lll}
        \hline
        \multirow{2}{*}{Model} & \multicolumn{3}{c}{\pfour}   && \multicolumn{3}{c}{\peight}     \\\cline{2-4} \cline{6-8}
                                & MAE  & RMSE  & MAPE           && MAE  & RMSE  & MAPE               \\ \hline
 
        w/o Mask                   & 19.32 & 31.23& 12.71         && 15.86 & 24.98 & 10.30          \\        
        w/o GVF                     & 19.23 &  31.35 & 12.60         && 15.44 &  24.81 & 10.74          \\       w/o  SVF                & 19.20 &  31.31 & 12.69         && 15.36 &  24.64 & 9.87          \\      
        w/o $\bm{A}_{\rm{S}}$      & 19.25 &  31.06 &  12.76         && 15.68 &  25.05 & 9.95          \\               
        \hline 
        \textbf{\name}           & \textbf{18.97}    & \textbf{30.99}    & \textbf{12.57}  && \textbf{15.12} & \textbf{24.56} &  \textbf{8.89} \\
        \hline
    \end{tabular}}
    
    \label{tab:ablation}
\end{table}

\subsection{Performance Comparison}

 \noindent We compare \name with the following baselines belonging to 3 classes. (1) classic time series forecasting methods: HA, ARIMA with Kalman filter, VAR, and FC-LSTM.  (2) STGNN-based models: STGCN, WaveNet, ASTGCN, MSTGCN, LSGCN, STSGCN, and STFGNN. (3) {NDE-based models}: STGODE and STG-NCDE. 

Table~\ref{tab:main_exp} and Table~\ref{tab:main_exp_2} show the average forecasting performances over 12 horizons of \name and baselines on 6 real-world traffic datasets. 
We conduct statistical tests to verify if the performance differences are statistically significant. CEGNCDE is statistically superior to the compared models according to pairwise T-test at a 95\% significance level.
The following phenomena can be observed:
\begin{itemize}

\item STGNN-based models achieve better results than classic time series forecasting methods, which only take temporal dependencies into consideration and ignore spatial dependencies, whereas STGNN-based models can take advantage of both temporal dependencies and spatial dependencies.

\item Although STGODE is a NDE-based model, it simply replaces the residual network with NODE. Like STGNN-based models, STGODE cannot capture the continuous temporal dependencies and spatial dependencies, so its performance is similar to that of the SOTA STGNN-based models. Due to the abilities to capture continuous temporal dependencies and spatial dependencies, STG-NCDE has an average superior performance than STGNN-based models.

\item \name (ours) outperforms all baselines on 6 real-world traffic datasets, achieving a SOTA traffic  forecasting performance. Compared to the best baseline model, \name yields an average \upMAE relative MAE reduction, \upRMSE relative RMSE reduction, and \upMAPE relative MAPE reduction. The increase is significant in this well studied area, which indicates the benefits of modeling spatial dependencies continuously evolving over time.
\end{itemize}

\subsection{Ablation Study}
 \noindent To further investigate the effectiveness of different parts in \name, we compare \name with the following variants on \pfour and \peight : \textbf{w/o Mask}, \textbf{w/o GVF}, \textbf{w/o SVF}, and \textbf{w/o $\bm{A}_{\rm{S}}$} remove two mask matrices $\bm{M}^{\rm{geo}}$ and $\bm{M}^{\rm{sem}}$, GVF, SVF, and the stable static adjacency matrix $\bm{A}_{\rm{S}}$, respectively. 

We conduct statistical tests to verify if the performance differences are statistically significant. CEGNCDE is statistically superior to all variants according to pairwise T-test at a 95\% significance level. Table~\ref{tab:ablation} shows the comparison of these variants. Based on the results, the following phenomena can be observed:
\begin{itemize}

    
    \item \name leads to a large performance improvement over \textbf{w/o Mask}, highlighting the value of using the mask matrices to identify the significant node pairs.
    \item \textbf{w/o GVF} and \textbf{w/o SVF} perform worse than \name, indicating that both local and global spatial dependencies are significant for traffic forecasting.
    \item \textbf{w/o $\bm{A}_{\rm{S}}$} performs worse than \name, indicating the indispensable role of the relatively stable component in capturing spatial dependencies.
    
\end{itemize}


\begin{figure}[!ht]
\centering
\subfigure[Node 5, 12 and 17 on \peight] {\includegraphics[width=1\linewidth]{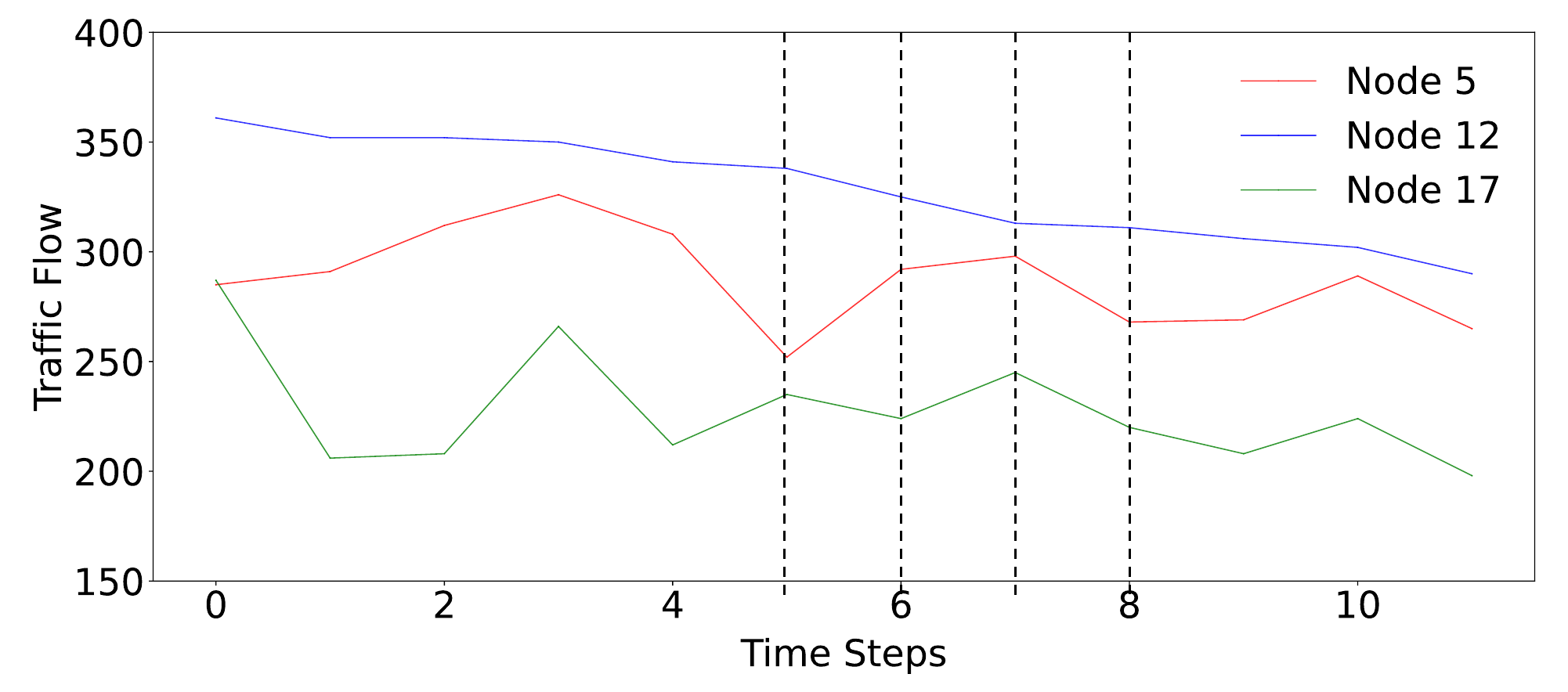}} 
\subfigure[Time Step 5]{\includegraphics[width=0.47\linewidth]{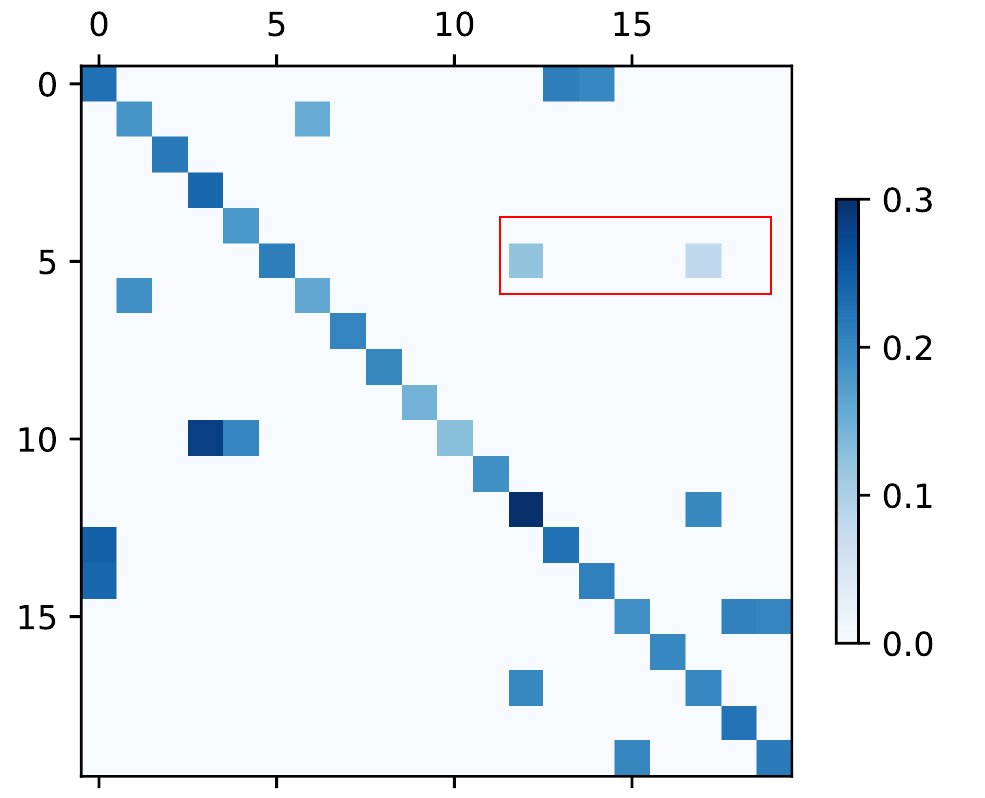}}\hspace{5mm}
\subfigure[Time Step 6]{\includegraphics[width=0.4\linewidth]{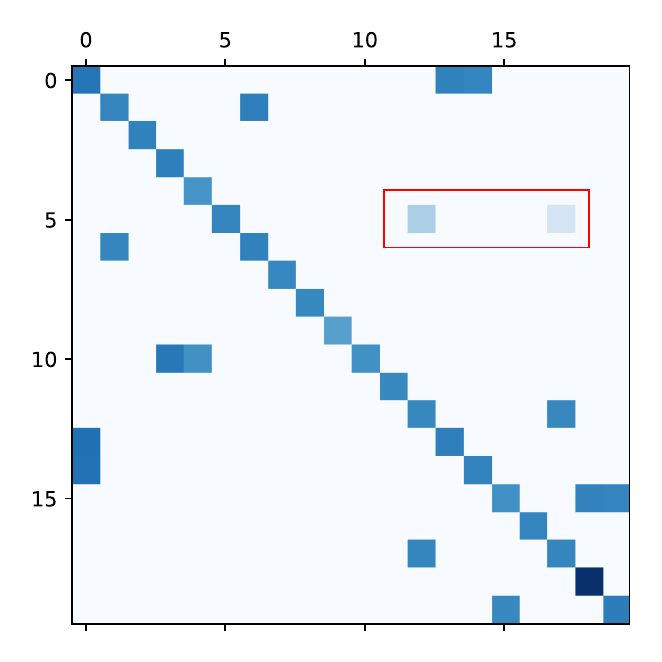}}
\subfigure[Time Step 7]{\includegraphics[width=0.4\linewidth]{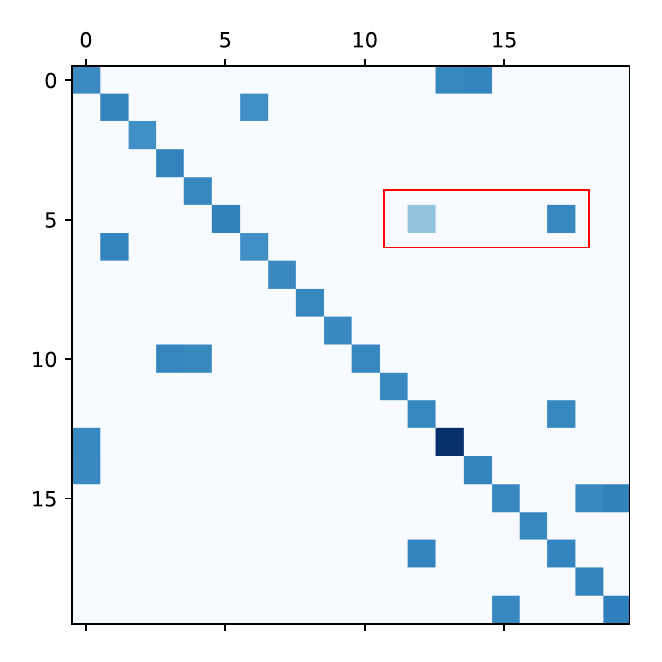}}\hspace{11mm}
\subfigure[Time Step 8]{\includegraphics[width=0.4\linewidth]{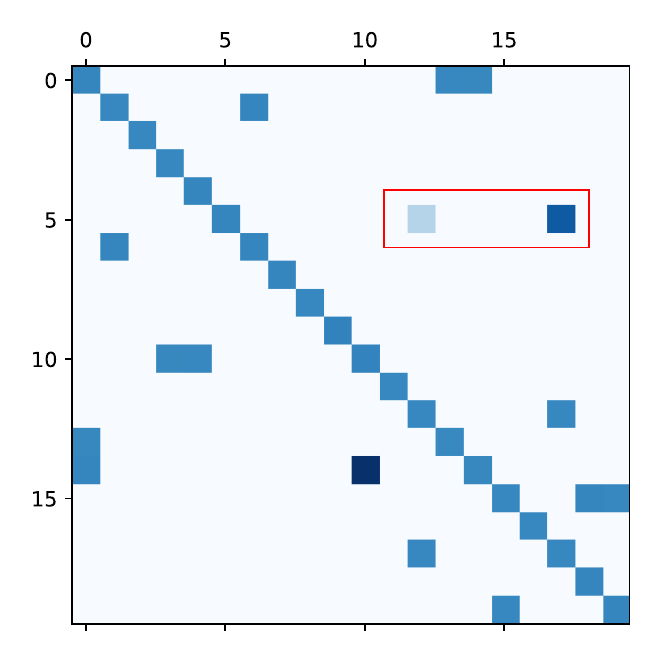}}
\caption{Case Study of CEG.} 
\label{fig:case}  
\end{figure}
        
 

\subsection{Case Study}
 \noindent To further illustrate the modeling of continuous spatial dependencies over time, we conduct case studies on \peight. We randomly select a period traffic data of three nodes, i.e., node 5, 12, and 17, and plot their ground truth, as shown in Figure~\ref{fig:case}(a). Both node 12 and 17 are the semantic neighbors of node 5 (semantic neighbor means that node pair have a high similarity by calculating the DTW distance). We visualize the CEG with semantic mask containing the first 20 nodes at time step 5, 6, 7, and 8, as shown in Figure~\ref{fig:case}(b)(c)(d)(e), respectively. The red windows in CEG contain the spatial dependencies between node 5 and 12, as well as node 5 and 17. At time step 5 and 6, the trend of node 5 and 17 is completely opposite, and the corresponding positions in CEG have extremely light colors. At time step 7 and 8, the trend of node 5 and 17 is more similar, and the corresponding positions in CEG are darker in color. In addition, we observed that the trend similarity between node 5 and 12 is always very small, and the corresponding position color in CEG is always very light. These all indicate that our proposed CEG can reflect the evolution of spatial dependencies over time.

\section{Conclusions}
 \noindent In this paper, we propose CEGNCDE to capture continuous spatial-temporal dependencies. Specifically, a continuously evolving graph generator is introduced to generate the spatial dependencies graph that continuously evolves over time from discrete historical observations. Then, a graph neural controlled differential equations framework is introduced to capture continuous temporal dependencies and spatial dependencies over time simultaneously. Comprehensive experiments are conducted on 6 real-world traffic datasets and the results demonstrate the superiority of \name.

\bibliographystyle{named}
\bibliography{ijcai24}

\end{document}